\documentclass[conference]{IEEEtran}
\IEEEoverridecommandlockouts

\usepackage{cite}
\usepackage{amsmath,amssymb,amsfonts}
\usepackage{algorithmic}
\usepackage{textcomp}
\usepackage{xcolor}
\usepackage{graphicx}
\usepackage{utfsym}
\usepackage{subfigure}
\usepackage{wrapfig}
\usepackage{amsfonts}
\usepackage{color}
\usepackage{CJKutf8}
\usepackage{amsmath}
\usepackage{url}
\newtheorem{definition}{Definition}

\def\BibTeX{{\rm B\kern-.05em{\sc i\kern-.025em b}\kern-.08em
    T\kern-.1667em\lower.7ex\hbox{E}\kern-.125emX}}
\begin{document}

\title{Reliable Imputed-Sample Assisted Vertical Federated Learning\\
}

\author{\IEEEauthorblockN{Yaopei Zeng}
\IEEEauthorblockA{\textit{Shenzhen Research Institute of Big Data} \\
\textit{The Chinese University of Hong Kong, Shen Zhen}\\
Shenzhen, China \\
zengyaopei@gmail.com
}

\and
\IEEEauthorblockN{Li Liu}
\IEEEauthorblockA{
\textit{The Hong Kong University of Science and Technology (Guangzhou)}\\
Guangzhou, China \\
avrillliu@hkust-gz.edu.cn}
\and
\IEEEauthorblockN{Lei Liu}
\IEEEauthorblockA{\textit{Shenzhen Research Institute of Big Data} \\
\textit{The Chinese University of Hong Kong, Shenzhen}\\
Shenzhen, China \\
liulei1497@gmail.com}
\and

\IEEEauthorblockN{Shaoguo Liu}
\IEEEauthorblockA{
\textit{Alibaba Group}\\
Beijing, China \\
shaoguo.lsg@alibaba-inc.com}
\and
\IEEEauthorblockN{Hongjian Dou}
\IEEEauthorblockA{
\textit{Alibaba Group}\\
Beijing, China \\
hongjian.dhj@alibaba-inc.com}
}

\author{\IEEEauthorblockN{Yaopei Zeng$^{1}$, Lei Liu$^{2}$, Shaoguo Liu$^{3}$, Hongjian Dou$^{3}$, Baoyuan Wu$^{2}$, Li~Liu$^{4,*}$\thanks{*Corresponding author: avrillliu@hkust-gz.edu.cn.}}

\IEEEauthorblockA{$^{1}$\textit{The Pennsylvania State University, State College, USA}\\
$^{2}$\textit{The Chinese University of Hong Kong, Shenzhen}~~~~~
$^{3}$\textit{Alibaba Group, Beijing, China}\\
$^{4}$\textit{Hong Kong University of Science and Technology (Guangzhou), Guangzhou, China}
} 
}

\maketitle

\begin{abstract}
Vertical Federated Learning (VFL) is a well-known FL variant that enables multiple parties to collaboratively train a model without sharing their raw data. Existing VFL approaches focus on overlapping samples among different parties, while their performance is constrained by the limited number of these samples, leaving numerous non-overlapping samples unexplored. 
Some previous work has explored techniques for imputing missing values in samples, but often without adequate attention to the quality of the imputed samples.
To address this issue, we propose a \textit{R}eliable \textit{I}mputed-\textit{S}ample \textit{A}ssisted \textit{(RISA)} VFL framework to effectively exploit non-overlapping samples by selecting reliable imputed samples for training VFL models. Specifically, after imputing non-overlapping samples, we introduce evidence theory to estimate the uncertainty of imputed samples, and only samples with low uncertainty are selected. In this way, high-quality non-overlapping samples are utilized to improve VFL model. Experiments on two widely used datasets demonstrate the significant performance gains achieved by the RISA, especially with the limited overlapping samples, \textit{e.g.}, a 48\% accuracy gain on CIFAR-10 with only 1\% overlapping samples.
\end{abstract}

\begin{IEEEkeywords}
Vertical Federated Learning, Evidence Theory
\end{IEEEkeywords}

\section{Introduction}
\label{sec:intro}

\begin{figure}[t]
\centering
\includegraphics[width=0.95\linewidth]{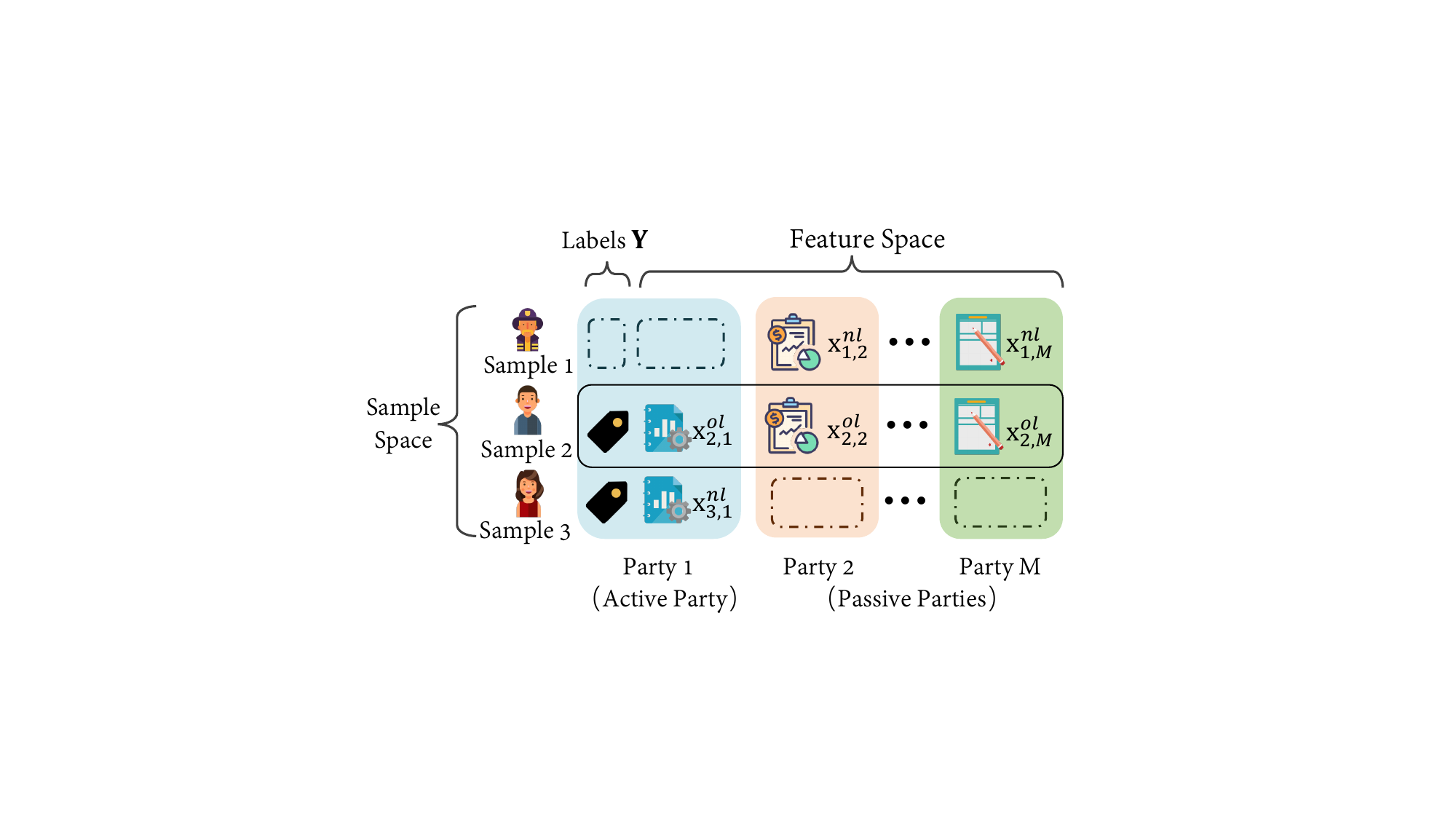}
\caption{Illustration of the virtual dataset in the context of VFL involving $M$ parties. Each party holds a vertical proportion of this dataset. The party with the labels is known as the active party, whereas the other parties are referred to as passive. Sample 2 is shared among parties, named overlapping samples. Samples 1 and 3 are non-overlapping since they have missing attributes and labels. The absent attributes and labels are depicted as hollow rectangles.}
\label{vfl_data}
\end{figure}

With increasing concerns about data privacy, Federated Learning (FL) has emerged as a privacy-preserving learning approach that allows multiple data collectors to collaboratively train models without sharing their raw data \cite{fl, flsurvey,  li2021survey, kairouz2021advances, zeng2023global}. Vertical Federated Learning (VFL), as a widely studied FL variant, mainly focuses on different attribute spaces of the overlapping data owned by multiple parties \cite{vfl, feng2020multi, wei2022vertical, yang2019parallel, dai2021vertical, wei2023fedads}. As illustrated in Fig. \ref{vfl_data}, each party contributes unique attributes\footnote{In this work, ``attributes" refers to the raw value of a sample before encoding, while ``features" refer to the representations produced by the neural network after encoding.} to form a comprehensive attribute space for the overlapping data (e.g., Sample $2$). The party that provides the training labels is referred to as the active party, while the others are termed passive parties. Some samples, such as Sample $1$ and Sample $3$, are non-overlapping due to the absence of certain attributes. Additionally, in the active party, non-overlapping samples also lack labels (e.g., Sample $1$).


Previous approaches \cite{kang2022fedcvt, sfhtl, sun2023communication, zou2022fedmc, fedda} have been proposed to address the issue of limited overlapping data in VFL. The main solution \cite{sun2023communication, zou2022fedmc} is to impute missing attributes for non-overlapping data relying on predefined linear functions. For example, Kang \textit{ et al}. \cite{kang2022fedcvt} and Sun \textit{ et al}. \cite{sun2023communication} estimate the missing features by leveraging the scaled dot-product attention function to calculate the similarity between overlapping and non-overlapping samples. However, their effectiveness remains constrained due to the feature generation quality, which is affected by the limited number of overlapping samples.


To address the aforementioned challenge, we propose a novel framework called the \textbf{R}eliable \textbf{I}mputed-\textbf{S}ample \textbf{A}ssisted (RISA) VFL model to evaluate the quality of non-overlapping samples with uncertainty estimation. RISA imputes non-overlapping samples by completing missing attributes with mean imputation \cite{JAMSHIDIAN200721} and estimating labels through self-training \cite{amini2022self}. To evaluate the quality of imputed non-overlapping samples, RISA incorporates evidence learning \cite{edl}, which leverages features from different parties at the evidence level, rather than at the feature or output level, to provide a reliable uncertainty estimation for imputed samples within each party. This approach allows RISA to assess the quality of both raw features and imputed features from each party, facilitating weighted dynamic collaboration among parties and the selection of non-overlapping samples for VFL training. 
By integrating these innovative components, RISA effectively utilizes reliable non-overlapping samples to improve model performance in VFL. Experimental evaluations conducted on various datasets with different proportions of overlapping samples demonstrate the superiority of RISA over state-of-the-art (SOTA) methods, particularly when the overlap is limited.

In summary, this work has the following three main contributions: 
\textbf{(1)} \textbf{Development of the RISA Framework}: To enhance VFL, the proposed RISA framework includes not only non-overlapping sample imputation, but also the quality assessment of imputed samples. This framework allows for the effective utilization of non-overlapping data, which is typically neglected in existing VFL models.
\textbf{(2)} \textbf{Uncertainty Estimation with Evidence Theory}: We propose a novel strategy to assess the quality of samples, that is estimate the uncertainty of samples based on the evidence theory. By incorporating this technique, only reliable imputed samples are selected for VFL training, ensuring that the noise from inaccurate imputations is minimized, thereby improving overall model performance.
\textbf{(3)} \textbf{Demonstrated Performance Gains}: Extensive experiments on two widely used datasets (CIFAR-10 and Criteo) show that RISA significantly improves the accuracy of VFL models, particularly when there is a limited overlap between the datasets of participating parties. For instance, a 48\% accuracy improvement is achieved on CIFAR-10 with only 1\% overlapping samples.

\begin{figure*}
    \centering
     \includegraphics[width=0.8\linewidth]{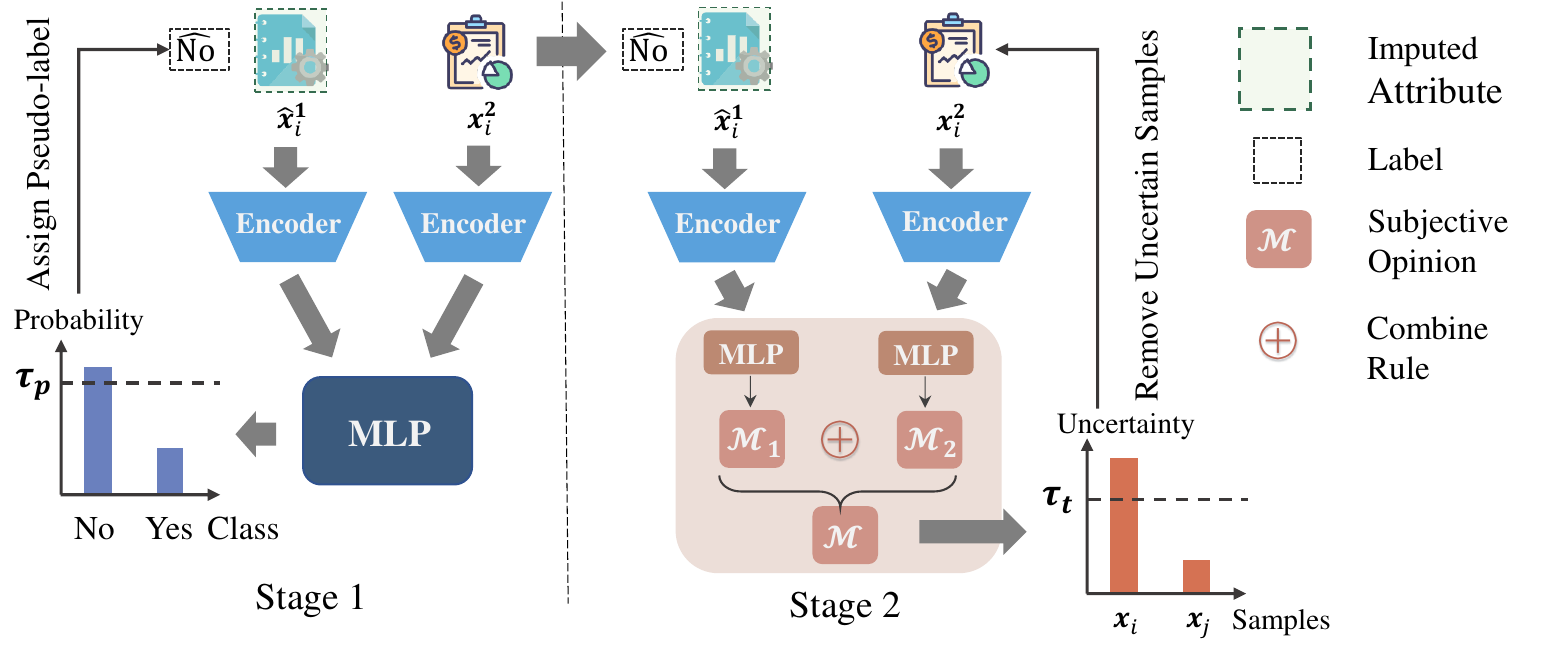}
    \caption{The pipeline of the proposed RISA framework consists of three stages. In \textbf{Stage 1}, RISA adopts the mean imputation to impute attributes and the self-training strategy to assign pseudo-labels for non-overlapping samples in the passive party (\textit{i.e.}, Party 2). In \textbf{Stage 2}, RISA projects the features into evidence vectors with MLPs. Subjective opinions are formed from each party based on DST theory and fused using the combination rule described in Definition \ref{combination} for the final prediction.}
    \label{framework}
\end{figure*}

\section{Methodology}
\label{sec:pagestyle}
\subsection{Problem Formulation}

We consider a typical VFL setting involving $M$ parties, where each party possesses a private dataset. Party $m$ owns the dataset $\mathbf{D}^m=\{\mathbf{X}^m_i\}^{n^m}_{i=1}$, with $n^m$ representing the size of the dataset. 
Without loss of generality, we assume that the dataset of the first party contains the ground-truth labels, denoted as $\mathbf{Y} \in \mathbb{R}^{n_1 \times 1}$. As depicted in Fig. \ref{vfl_data}, $\mathbf{D}^m$ of $m$-th party has overlapping samples $\mathbf{O}^m$ and non-overlapping samples $\mathbf{N}^m$, where $n^{m}_{o}$ and $n^{m}_{non}$ denote the number of overlapping and non-overlapping samples, respectively. 



\subsection{Reliable Imputed-Sample Assisted Framework}

As shown in Fig. \ref{framework}, RISA consists of two important stages: (1) completing non-overlapping samples via mean imputation and self-training, and (2) evidential VFL for uncertainty estimation based on Evidence Theory, achieving a dynamic weighted collaboration. 


\subsection{Imputing Non-overlapping Samples}
\label{semisec}
To handle non-overlapping samples in VFL, we impute the missing attributes by leveraging the mean values derived from the overlapping samples within each party. Specifically, we calculate the mean value for the overlapping samples as $\overline{\mathbf{x}}^m=\frac{\sum \mathbf{O}^m}{n_o^m}$ for each party $m$ where $m \in\{1, \ldots, M\}$. Then, for a non-overlapping sample $\mathbf{x}_i$ associated with party $l$, the imputed sample $\hat{\mathbf{x}}_i$ is constructed as follows:
$$
\hat{\mathbf{x}}_i=\operatorname{concat}\left(\mathbf{x}_i^l, \overline{\mathbf{x}}^1, \overline{\mathbf{x}}^2, \ldots, \overline{\mathbf{x}}^{l-1}, \overline{\mathbf{x}}^{l+1}, \ldots, \overline{\mathbf{x}}^M\right),
$$
where $\mathbf{x}_i^l$ represents the observed attributes of the non-overlapping sample held by party $l$. Mean imputation is a more efficient and privacy-preserving method compared to the imputation strategies used in prior studies \cite{kang2022fedcvt, sfhtl, sun2023communication}, since it only necessitates the exchange of statistical data among parties. Despite its simplicity, our experiment results demonstrate that it offers a competitive performance improvement.

The non-overlapping data from passive parties lack ground-truth labels. To effectively utilize these samples, we propose a self-training strategy that facilitates automatic pseudo-label assignment \cite{tur2005combining, lee2013pseudo}. This approach involves a vertical self-training model, denoted as $f_{ST}$ denote the self-training mode, which is parameterized by $\theta$ and optimized on overlapping samples using the cross-entropy loss:
\begin{equation}
    \mathcal{L}_{CE} = \sum_{k=1}^K y_{ik}log(f_{ST}(\theta; \mathbf{x}_i)), 
\end{equation}
where $y_{ik}$ represents the ground-truth label for overlapping sample $i$ for class $k$. Using $f_{ST}$, a pseudo-label for a non-overlapping sample $\mathbf{n}_j$ is assigned to the $j$-th class if the predicted probability $p_{jk}$ meets or exceeds a predefined threshold $\tau_p$, where $p_{jk}$ is derived from the output vector $\mathbf{p}_{j} = f_{ST}(\theta; \mathbf{n}_j)$. Through experimentation, we have observed two issues: (1) A higher threshold $\tau_p$ significantly reduces the frequency of pseudo-label assignments, resulting in many non-overlapping samples remaining unlabeled. (2) In contrast, a lower threshold tends to introduce noisy labels, which affects the performance of the model. To address these challenges, we introduce an evidential VFL method based on evidence theory \cite{yager1994advances}, which is further elaborated in the following Section \ref{tru}.

\subsection{Evidential VFL via Evidence Theory}
\label{tru}
Although the process in Section \ref{semisec} can incorporate non-overlapping samples into the VFL, it inevitably introduces attribute and label noise, especially when overlapping samples are scarce. In response to these challenges, and inspired by evidence theory \cite{yager1994advances, edl, yu2024anedl}, we propose the Evidential Vertical Federated Learning (EVFL) algorithm. This innovative approach quantifies the uncertainty associated with each sample, facilitating a thorough uncertainty evaluation of the imputed non-overlapping samples.

\textbf{Evidential Deep Learning}.
The evidential deep learning typically utilizes empster-Shafer Theory of Evidence (DST) \cite{shafer1992dempster} and Subjective Logic (SL) \cite{sl} for uncertainty estimation \cite{edl, bao2021evidential, zhao2023open, park2022active, aguilar2023continual, xia2024uncertainty}. In this setup, for a classification problem encompassing $K$ classes, the softmax function typically used in the final layer of the classification model in party $m$ is substituted with an activation function. This adjustment enables the model output $K$ non-negative values $\mathbf{e}^m=[e^m_1, \ldots, e^m_K]$ that correspond to the $K$ classes for each sample, which is the evidence value in evidential deep learning. According to SL, these evidence values are utilized to compute the parameters of the Dirichlet distribution $\mathbf{\alpha}^m=\left[\alpha_1^m, \ldots, \alpha_K^m\right]$ where $\alpha_k^m=e_k^m+a_k$ with $a_k$ as the class-specific prior, typically assumed to be 1 in line with prior research \cite{han2022trusted, han2020trusted}.
Then, $K$ belief masses $\mathbf{b}^m=[b^m_1, \ldots, b^m_K]$ and a single overall uncertainty mass $u^m$ are computed as:
$$
b_k^m=\frac{e_k^m}{S^m}=\frac{\alpha_k^m-1}{S^m} \quad \text {and} \quad u^m=\frac{K}{S^m},
$$
where $S^m=\sum_{i=1}^K\left(e_i^m+1\right)=\sum_{i=1}^K \alpha_i^m$ is the Dirichlet strength. 
The belief mass $b^m_k$ represents the possibility that the sample belongs to class $k$ and the uncertainty $u$
These values collectively form an opinion $\mathcal{M}^{m}$. It is imperative to note that both $u^m \geq 0$ and $b^m_k \geq 0$ for all $k=1, \ldots, K$, and their sum is equal to one, as represented by the equation:

\begin{equation}
    u^m+\sum_{k=1}^K b^m_k=1.
\end{equation}
Therefore, the value of uncertainty is inversely proportional to the total evidence provided by the model, indicating that greater evidence corresponds to reduced uncertainty in the classification outcomes.


\textbf{Evidential Vertical Federated Learning (EVFL)}.
In EVFL, as depicted in Fig. \ref{framework}, each party forms an opinion $M^m$ based on the features of a given sample. These individual opinions must be aggregated to derive the final prediction. Traditionally, Dempster’s combination rule \cite{shafer1992dempster} is employed for this aggregation \cite{han2022trusted, bao2021evidential}. However, this rule can produce counterintuitive outcomes in cases of significant conflict among opinions \cite{conflict1, conflict2, conflict3}, often arising from the ambiguous features of non-overlapping samples in VFL. To address this issue, we have developed a modification termed the Reduced Yager's Combination Rule, based on the original Yager's Combination Rule \cite{yager1987dempster, yager1994advances}. This rule is designed to measure and effectively manage the conflict among opinions, thereby enhancing the integration process during the training of the participating parties.

\begin{definition} (\textbf{Reduced Yager's Combination Rule})
\label{combination}
Given two sets of masses $\mathcal{M}^1=\left\{\left\{b_k^1\right\}_{k=1}^K, u^1\right\}$ and $\mathcal{M}^2=\left\{\left\{b_k^2\right\}_{k=1}^K, u^2\right\}$, the combined mass set $\mathcal{M}=\{\{b_k\}_{k=1}^K,u\}$ is derived by the operation $\mathcal{M}=\mathcal{M}^1 \oplus \mathcal{M}^2$.
The combination is computed as follows:
\begin{equation}
b_k=b_k^1 b_k^2+b_k^1 u^2+b_k^2 u^1, u=u^1 u^2 + C\text {, }
\end{equation}
where $C=\sum_{i \neq j} b_i^1 b_j^2$ quantifies the conflict between the two sets of belief masses.
\end{definition}

In a scenario involving features from $M$ distinct parties, $M$ opinions are formed and subsequently fused using the Reduced Yager’s Rule of combination:
\begin{equation}
\mathcal{M}=\mathcal{M}^1 \oplus \mathcal{M}^2 \oplus \cdots \mathcal{M}^M.
\end{equation}

The overall strength of the Dirichlet distribution $S$ and parameters of Dirichlet distribution for class $k$ are determined as follows:
\begin{equation}
    S=\frac{K}{u} \text { and } \alpha_k=b_k \times S+1,
\end{equation}
where $u$ denotes the combined uncertainty, and $b_k$ represents the combined belief mass for class $k$.

This combination rule implies that opinions with higher uncertainty contribute less to the overall belief mass. Moreover, the final aggregated opinion exhibits high uncertainty under two conditions: 1) when all contributing opinions are marked by significant uncertainty, and 2) when there is strong conflict among the opinions.
For the prediction purpose,  the probability for class $k$ is estimated by the mean of the corresponding Dirichlet distribution: $ \hat{p}_k=\frac{\alpha_k}{S}$. 

The final loss function in EVFL is:
\begin{equation}
\begin{aligned}
    \mathcal{L}(\boldsymbol{\alpha}_i) 
    =\sum_{k=1}^K y_{i k}\left(\log \left(S_i\right)-\log \left(\alpha_{i k}\right)\right),
\end{aligned}
\end{equation}
where $y_{ik}$ is the label of sample $i$ for class $k$. 

During training, we adopt a strategy to exclude samples that display high uncertainty, which may arise from incorrect pseudo-labels or ambiguously generated features. Specifically, the average uncertainty of each sample is calculated at intervals of $E$ epochs. Samples exceeding a specified uncertainty threshold are then removed from the training set in subsequent epochs. Generally, the certainty of the predictions on the training samples increases as the training progresses, with the uncertainties typically decreasing. To accommodate this pattern, we dynamically adjust the uncertainty threshold during the training process.
The uncertainty threshold $\tau_t$ at epoch $t$ is defined as $\tau_t = \tau_0 ^{\frac{t}{T}}$, where $T$ is the total number of training epochs. The hyper-parameter $\tau_0$ is set as the final uncertainty threshold. As training progresses, the threshold $\tau_t$ progressively decreases from 1 to $\tau_0$, aligning with the increasing certainty of model predictions.

\section{Experiments}
\label{exp}
\subsection{Experimental Setting}
\label{setting}

\textbf{Datasets and VFL Setup.} We conduct experiments on two datasets, namely CIFAR-10 \cite{cifar} and Criteo\cite{criteo}.

(1) \textbf{CIFAR-10} dataset consists of 60,000 color images with dimensions of 32$\times$32$\times$3 pixels, categorized into 10 classes. To simulate the two-party VFL scenario, we adopt the partitioning approach utilized in \cite{kang2022fedcvt, liu2021rvfr}, where each CIFAR-10 image is divided into two parts, each with 32$\times$16$\times$3 pixels. 

(2) \textbf{Criteo} dataset is a public benchmark dataset for click-through rate (CTR) prediction. It has 13 numerical features and 26 categorical features. We used 9,000,000 samples for training and 1,000,000 samples for testing. The test metric is the area under the ROC curve (AUC). The feature fields are manually split into two parties, where Party 1 has all numerical fields and Party 2 has categorical fields.

\textbf{Baselines.} Our framework is compared with the following baselines: (1) Local: Utilizes only the features from Party 1. (2) VFL: Employs a model trained on overlapping samples, processed through supervised VFL,
(3) Local + VFL: Non-overlapping samples from Party 1 are imputed with zeros and then treated as if they were overlapping samples.
(4) Random Match: Non-overlapping samples are randomly paired. 
(5) SSL + VFL: Integrates both overlapping and non-overlapping samples in training, utilizing SimCLR \cite{simclr} for pre-training, followed by supervised VFL processing.
(6) SFHTL \cite{sfhtl}: A transfer federated learning framework.
(7) FedCVT \cite{kang2022fedcvt}: A semi-supervised VFL framework.
(8) Few-shot VFL \cite{sun2023communication}: A communication-efficient VFL framework.

\begin{table}[t]
\centering
\caption{Test accuracy (\%) of different methods on CIFAR-10 with various overlapping proportions.}
\resizebox{0.95\linewidth}{!}{
\begin{tabular}{l|c|c|c|c|c|c}
\hline
Method    & 1\%              & 10\%             & 20\%            & 30\%              & 40\%              & 50\%\\ \hline
Local     & 81.78   & 81.78   & 81.78  & 81.78   & 81.78    & 81.78 \\
VFL       & 38.58   & 70.47   & 78.11  & 81.78     & 83.97    & 85.45 \\
Local + VFL & 85.36 & 86.11     & 86.14     & 87.85 & 89.47 & 89.39 \\
Random Match& 83.53 & 84.08    & 84.96     & 85.48     & 85.87    & 86.41 \\ 
SSL + VFL   & 74.11 & 75.42 & 75.98 & 76.08 & 76.61 & 76.33  \\ 
SFHTL \cite{sfhtl}  & 85.47    & 85.89     & 86.05     & 86.56    & 86.92 & 87.76 \\ 
FedCVT \cite{kang2022fedcvt}  & 85.93    & 86.85     & 86.99     & 87.23    & 87.77& 88.38 \\ 
Few-shot VFL \cite{sun2023communication}  & 82.17    & 86.13     & 86.43     & 86.55    & 87.02 & 87.12 \\ 
RISA (Ours) &\textbf{87.62}  & \textbf{88.65} & \textbf{88.97} & \textbf{89.03} & \textbf{89.55} & \textbf{89.82}\\ \hline
\end{tabular}}
\label{cifar}
\end{table}

\begin{table}[t]
\centering
\caption{Test AUC (\%) of different methods on Criteo with various overlapping proportions.}
\resizebox{0.95\linewidth}{!}{
\begin{tabular}{l|c|c|c|c|c|c}
\hline
Method    & 1\%     & 10\%      & 20\%      & 30\%      & 40\%      & 50\% \\ \hline
Local     & 68.03 & 68.03 & 68.03 & 68.03 & 68.03 & 68.03 \\
VFL       & 67.96 & 70.43 & 73.99 & 74.56 & 74.89 & 75.43 \\
Local + VFL & 73.46 & 76.30 & 76.63 & 77.10 & 77.05 & 77.27  \\
Random Match& 73.43 & 73.58 & 74.42 & 74.67 & 75.18 & 75.84 \\ 
SSL + VFL   & 74.11 & 75.42 & 75.98 & 76.08 & 76.61 & 76.33  \\ 
SFHTL \cite{sfhtl}  & 74.28 & 75.48 & 75.81 & 76.02 & 76.25 & 76.08 \\ 
FedCVT \cite{kang2022fedcvt}  & 74.37  & 74.94  & 75.73  & 75.65 & 76.03 & 76.51 \\ 
Few-shot VFL \cite{sun2023communication}  & 72.85   & 73.79    & 74.75     & 74.96    & 75.72 & 75.94 \\ 
RISA (Ours) & \textbf{75.67} & \textbf{77.36} & \textbf{77.62} & \textbf{77.81} & \textbf{77.87} & \textbf{77.89} \\ \hline
\end{tabular}}
\label{cretio}
\end{table}
\subsection{Results Analysis}
\label{exp_result}
 
As demonstrated in Tables \ref{cifar} and \ref{cretio}, our model consistently outperforms other methods across datasets with varied data types. Notably, the performance advantage of our model becomes more pronounced as the number of overlapping samples decreases. Significantly, the performance advantage of our model intensifies as the number of overlapping samples decreases. For instance, when the proportion of overlapping samples is reduced to 1\%, our framework achieves a 48\% increase in accuracy on the CIFAR-10 dataset.


\begin{table}[h]
\centering
\caption{Component analysis of RISA.}
\resizebox{0.7\linewidth}{!}{
\begin{tabular}{c|c|c|c|c}
\hline
IMP        & ST           & EVFL            & CIFAR-10        & Cretio \\ \hline
\usym{2613} & \usym{2613}  & \usym{2613}     & 70.47          & 70.43  \\ \hline
\checkmark  & \usym{2613}  & \usym{2613}     & 82.19          & 71.17  \\ \hline
\checkmark  & \checkmark   & \usym{2613}     & 85.01          & 74.45  \\ \hline
\checkmark  & \usym{2613}  & \checkmark      & 85.13          & 74.38  \\ \hline
\checkmark  & \checkmark   & \checkmark      & \textbf{88.65} & \textbf{77.36}  \\ 
\hline
\end{tabular}}
\label{component}
\end{table}

\subsection{Ablation Study}
Table \ref{component} illustrates the component analysis of our method, including impute attributes for non-overalpping samples (IMP),  self-training for assigning pseudo-labels to non-overalpping samples (ST), and evidential vertical federated learning strategy (EVFL). The experiments are conducted on CIFAR-10 and Cretio datasets with an overlapping proportion of 10\%. 
The results highlight the impact of each component on the overall performance. Specifically, the performance improves notably with imputation of non-overlapping samples and self-training, indicating the significant role of non-overlapping samples in improving the performance of VFL. Additionally, the inclusion of evidential deep learning also further contributes to improvement of the performance by decreasing the impact of imputation noise. 

\section{Conclusion}
 To overcome the challenge of limited overlapping samples in VFL, we propose a framework named RISA, which imputes missing attributes with mean values and predicts pseudo-labels through self-training for non-overlapping samples. Furthermore, RISA estimates the uncertainty of the predictions with evidence theory to learn a reliable model on the completed data set.
 Experimental results on two datasets demonstrate the significant performance improvements achieved by RISA, especially when the number of overlapping samples is limited.

\section*{Acknowledgment}
This work was supported by the National Natural Science Foundation of China (No. 62471420 and 62101351).

\vfill\pagebreak

\bibliographystyle{IEEEtran}
\bibliography{refs}

\end{document}